# Machine Learning Approaches to Clinical Risk Prediction: Multi-Scale Temporal Alignment in Electronic Health Records

Wei-Chen Chang*
University of Massachusetts Amherst
Amherst, USA

Lu Dai
University of California, Berkeley
Berkeley, USA

Ting Xu
University of Massachusetts Boston
Boston, USA

*Abstract-This study proposes a risk prediction method based on a Multi-Scale Temporal Alignment Network (MSTAN) to address the challenges of temporal irregularity, sampling interval differences, and multi-scale dynamic dependencies in Electronic Health Records (EHR). The method focuses on temporal feature modeling by introducing a learnable temporal alignment mechanism and a multi-scale convolutional feature extraction structure to jointly model long-term trends and short-term fluctuations in EHR sequences. At the input level, the model maps multi-source clinical features into a unified high-dimensional semantic space and employs temporal embedding and alignment modules to dynamically weight irregularly sampled data, reducing the impact of temporal distribution differences on model performance. The multi-scale feature extraction module then captures key patterns across different temporal granularities through multi-layer convolution and hierarchical fusion, achieving a fine-grained representation of patient states. Finally, an attention-based aggregation mechanism integrates global temporal dependencies to generate individual-level risk representations for disease risk prediction and health status assessment. Experiments conducted on publicly available EHR datasets show that the proposed model outperforms mainstream baselines in accuracy, recall, precision, and F1-Score, demonstrating the effectiveness and robustness of multi-scale temporal alignment in complex medical time-series analysis. This study provides a new solution for intelligent representation of high-dimensional asynchronous medical sequences and offers important technical support for EHR-driven clinical risk prediction.*

## I. INTRODUCTION

In recent years, with the rapid development of medical informatization and intelligent health management, Electronic Health Records (EHR) have become a crucial carrier of medical data. EHR systems play a central role in clinical diagnosis, health monitoring, and disease prediction. They systematically record various types of temporal information generated during patient visits, including vital signs, laboratory indicators, prescriptions, imaging examinations, and medical orders. These continuous, heterogeneous, and high-dimensional time-series data provide unprecedented opportunities for individualized health risk prediction. Compared with traditional static health archives, the dynamic temporal features contained in EHRs can reflect disease progression, treatment responses, and physiological trends, thus laying the data foundation for precision medicine and personalized healthcare[1].

However, EHR data present significant complexity and challenges. First, the frequency of visits, record lengths, and time intervals vary widely across patients, resulting in strong irregularity and non-uniformity in the time series. Second, EHR data include both structured and unstructured information, often exhibiting high dimensionality, noise, and severe missing values. Moreover, inconsistencies in recording standards and sampling frequencies across medical institutions further increase the difficulty of temporal alignment and feature fusion. These characteristics make it difficult for traditional time-series modeling methods to effectively capture multi-scale temporal dependencies and latent semantic relationships within EHR data, limiting their predictive performance in risk evaluation. Therefore, achieving multi-scale feature modeling and dynamic temporal alignment under non-uniform sampling has become a key research issue in medical artificial intelligence[2].

Risk prediction based on EHR sequences is not merely a technical problem but also a core requirement of clinical decision support systems. In applications such as early disease screening, chronic disease management, and risk warning, identifying high-risk individuals from massive EHR data is essential for reducing healthcare burdens, optimizing resource allocation, and improving survival rates. By deeply modeling temporal dependencies and cross-dimensional interactions in EHR sequences, key pathological patterns and physiological signal changes can be extracted from long-term patient records, enabling early detection and intervention. For patients with multiple coexisting diseases or complex disease trajectories, capturing multi-scale temporal features is especially critical for distinguishing between acute deterioration and stable states[3].

From a methodological perspective, traditional statistical or shallow learning-based risk models often assume uniform time intervals or ignore multi-scale structures, making it difficult to exploit the dynamic temporal nature of EHRs. Although recent deep learning models have advanced in feature representation [4-7], most focus on single-scale or fixed-step modeling, neglecting cross-granularity temporal features in healthcare data. The introduction of multi-scale temporal alignment networks addresses this limitation by incorporating learnable alignment mechanisms and hierarchical temporal modeling structures [8-11]. Such models can adaptively capture dynamic variations across different time scales, improving the unified representation of both short and long-term dependencies [12]. This design not only mitigates representational bias caused by irregular sampling but also enhances the model's sensitivity to key risk signals, providing more refined temporal modeling for EHR understanding.

In summary, developing a multi-scale temporal alignment network based on EHR sequences represents both a technological innovation for handling complex medical data and a significant step forward in intelligent health prediction and risk assessment. This research can deliver more interpretable and timely risk evaluations in clinical practice, supporting physicians with personalized recommendations and providing evidence for public health management and policymaking. More broadly, such models contribute to advancing the practical integration of artificial intelligence in healthcare, promoting the evolution of precision medicine, and driving the transformation from experience-based to data-driven medicine, thereby laying a solid theoretical and technical foundation for future intelligent healthcare systems.

## II. RELATED WORK

Deep learning has revolutionized the analysis of Electronic Health Records (EHR), offering powerful tools to capture complex temporal and semantic patterns within patient trajectories. Systematic reviews highlight that state-of-the-art deep models can effectively handle the irregularity and heterogeneity of EHR data, providing the methodological groundwork for intelligent health risk prediction and personalized medicine [13]. Building upon these foundations, hierarchical feature fusion and dynamic collaboration frameworks have been developed to further enhance feature representation. By combining multi-level feature extraction with adaptive integration, these architectures address the challenges of noisy, high-dimensional clinical signals and enable robust identification of clinically relevant patterns, even for small targets within large datasets [14].

To improve the generalizability of risk models across diverse disease types, neural networks have been widely applied for survival prediction in heterogeneous clinical cohorts. These methods demonstrate strong capacity to adaptively learn from varying medical scenarios, ensuring that prediction models remain effective in a broad range of real-world applications [15]. The success of deep learning in medical imaging, particularly with convolutional neural networks (CNNs), further illustrates the value of capturing spatial features and local context. CNN-based approaches are now standard for tasks such as cytopathology image classification, forming an essential part of many clinical decision support pipelines [16].

Beyond single-modal analysis, the integration of multimodal clinical data has become increasingly important. Deep learning-based multimodal fusion and semantic embedding techniques make it possible to combine structured and unstructured EHR sources, leading to richer, more informative representations that significantly improve risk prediction accuracy [17].

Taking a step further, contrastive learning frameworks for multimodal knowledge graph construction allow models to align information from disparate sources and reason over complex relationships in healthcare data. These frameworks provide enhanced support for data-analytical reasoning and facilitate more interpretable clinical inference [18]. Accurately modeling disease trajectories over time requires structure-aware temporal approaches. By explicitly representing temporal dependencies and patient state transitions, these methods enable fine-grained chronic disease progression prediction, which is key for proactive and personalized healthcare [19].

In addition, the emergence of retrieval-augmented generation within large language models introduces a new paradigm for medical knowledge fusion. By jointly modeling structured and unstructured information, these approaches can synthesize evidence across data types, empowering more comprehensive and context-aware clinical predictions [20]. For downstream clinical workflows, unified summarization and structuring of electronic medical records through deep learning and NLP methods further streamline data consumption. These approaches not only increase the interpretability of risk models but also make predictive insights more actionable for clinicians [21]. Finally, structure-learnable adapter fine-tuning has enabled the efficient transfer and adaptation of complex models to new clinical domains. By leveraging parameter-efficient learning strategies, these adapters facilitate continual improvement of risk prediction models as medical data landscapes evolve [22].

## III. METHOD

This study proposes a Multi-Scale Temporal Alignment Network (MSTAN) based on EHR sequences for risk characterization and prediction from irregularly sampled electronic health records. The model architecture is shown in Figure 1.

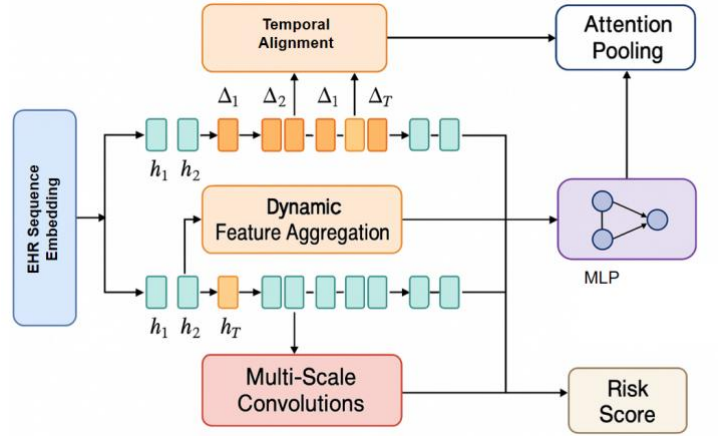

Figure 1. Overall model architecture diagram

The model consists of an input embedding layer, a multi-scale time alignment module, a dynamic feature aggregation layer, and a risk discrimination layer. First, for the original EHR sequence $X = \{x_1, x_2, ..., x_T\}$, where each $x_t \in R^d$ represents a multidimensional clinical feature at time t, the

model maps it to a high-dimensional representation space through a linear transformation:

$$h_t = W_e x_t + b_e \quad (1)$$

Where $W_e \in R^{d_h \times d}$ is the feature embedding matrix, and $b_e$ is the bias term. The embedded sequence $\{h_t\}$ retains the semantic structure of the original features and provides a continuous input basis for subsequent temporal modeling.

To address the uneven time intervals of EHR data, the model designs a learnable time alignment function to adaptively adjust the dependency strength between different time steps. For any two time points $t_i$ and $t_j$, the time difference between them is defined as $\Delta_{ij} = \|t_i - t_j\|$. The time alignment weight is modeled by a temporal attention function with a Gaussian kernel:

$$a_{ij} = \frac{\exp(-\frac{\Delta_{ij}^2}{2\sigma^2})}{\sum_{k=1}^{T} \exp(-\frac{\Delta_{ij}^2}{2\sigma^2})} \quad (2)$$

Where $\sigma$ is a learnable scale parameter that controls the rate of temporal decay. This mechanism allows the model to maintain key dependencies over long time spans, thereby achieving implicit alignment across different sampling frequencies. The temporal attention matrix $a$ corresponds to a temporal similarity map between sequences, effectively alleviating the heterogeneity of EHR data.

In the multi-scale feature modeling part, the model captures short-term and long-term dynamic features through the convolution of different scales and time window aggregation. Given a set of scales $S = \{s_1, s_2, ..., s_K\}$, each scale corresponds to a set of temporal convolution kernels $W^{(s)}$. The multi-scale convolution operation is defined as:

$$h_t^{(s)} = \sum_{k=-s}^{s} W_k^{(s)} \cdot h_{t+k} \quad (3)$$

By weighted fusion of multi-scale results, a unified multi-scale feature representation is obtained:

$$h = \sum_{s \in S} \beta_s \cdot h_t^{(s)} \quad (4)$$

$\beta_s$ is a learnable scale weight that reflects the contribution of different temporal granularities in risk characterization. This mechanism enables the model to capture pathological change patterns at different temporal levels, thereby improving the temporal understanding of asynchronous medical events.

In the final risk prediction stage, the model inputs the multi-scale time-aligned feature sequence into the global aggregation module and obtains the individual-level representation vector z through attention weighting:

$$z = \sum_{t=1}^{T} \gamma_t \widetilde{h}_t \quad where \quad \gamma_t = \frac{\exp(w^T \widetilde{h}_t)}{\sum_{i=1}^{T} \exp(w^T \widetilde{h}_t)} \quad (5)$$

Where A is a learnable attention vector that measures the importance of each time step in risk estimation. Finally, the risk score is calculated through a nonlinear mapping layer:

$$\hat{y} = \sigma(W_o z + b_o) \quad (6)$$

Where $W_o$ is the output weight matrix, $b_o$ is the bias term, and $\sigma(\cdot)$ represents the Sigmoid activation function. This structure achieves unified modeling of multi-scale semantics in continuous time and space, enabling the model to achieve stable risk representation and prediction capabilities in complex medical time series environments.

IV. EXPERIMENTAL RESULTS

A. Dataset

This study uses the publicly available MIMIC-III (Medical Information Mart for Intensive Care III) dataset. The dataset consists of electronic health records from the intensive care units (ICU) of a large medical institution, covering multimodal medical information from more than forty thousand hospitalized patients. The data include laboratory tests, medication prescriptions, vital signs, medical orders, clinical notes, and discharge summaries. The time span ranges from admission to discharge, providing a comprehensive view of patients' physiological changes and disease progression over different stages. Because it contains high-frequency and multi-scale temporal information, MIMIC-III has become one of the standard data sources for clinical time-series modeling and health risk prediction research.

During data preprocessing, the study selected structured variables closely related to risk assessment, such as blood pressure, heart rate, body temperature, respiratory rate, oxygen saturation, laboratory indicators, and medication records. The data were serialized according to timestamps. To mitigate inconsistencies in record length and sampling frequency across patients, the data were processed through temporal normalization, missing value imputation, and feature standardization. These steps ensured that the model could perform multi-scale feature modeling under a unified temporal framework. In addition, a time-window segmentation strategy was applied to preserve both long-term trends and short-term fluctuations, providing a stable foundation for subsequent temporal alignment and multi-scale learning.

The use of this dataset not only guarantees the reproducibility and verifiability of the study but also supports the model's generalization ability in complex medical environments. MIMIC-III covers a wide range of data types over an extended period, allowing the model to learn multimodal and cross-temporal associations effectively. Based on this dataset, the proposed multi-scale temporal alignment network can perform risk prediction under realistic clinical conditions, offering a generalizable framework for clinical decision support systems and personalized health management.

## B. Experimental Results

This paper first gives the results of the comparative experiment, as shown in Table 1.

Table1. Comparative experimental results

| Model | ACC | F1-Score | Precision | Recall |
|---|---|---|---|---|
| Transformer[23] | 0.871 | 0.864 | 0.881 | 0.848 |
| GCN[24] | 0.853 | 0.851 | 0.859 | 0.843 |
| GAT[25] | 0.878 | 0.872 | 0.888 | 0.861 |
| BERT[26] | 0.889 | 0.882 | 0.894 | 0.872 |
| Ours | 0.907 | 0.903 | 0.912 | 0.894 |

As shown in Table 1, the proposed multi-scale temporal alignment network surpasses all baseline models across evaluation metrics, confirming its effectiveness for EHR time-series modeling and clinical risk prediction. Compared with the traditional Transformer, it achieves notable gains in accuracy and F1-Score by leveraging multi-scale temporal modeling and dynamic alignment to better capture patient-state dependencies, especially in irregular and multi-granularity data. While graph-based models such as GCN and GAT achieve higher Precision but lower Recall due to limited temporal modeling, the proposed method's cross-scale alignment improves Recall by effectively identifying high-risk samples. Similarly, compared with BERT-based approaches, it attains superior F1-Score and Recall, demonstrating that static contextual encoding cannot fully represent temporal variability. By integrating time-weighted aggregation and scale fusion, the model captures both local fluctuations and global trends, ensuring robust predictions under multimodal inputs. Overall, these results validate that hierarchical temporal modeling combined with adaptive alignment yields unified and expressive representations of asynchronous, non-uniform medical data, significantly enhancing prediction accuracy, generalization, and clinical applicability. Figure 2 further presents the sensitivity analysis of attention temperature to Recall.

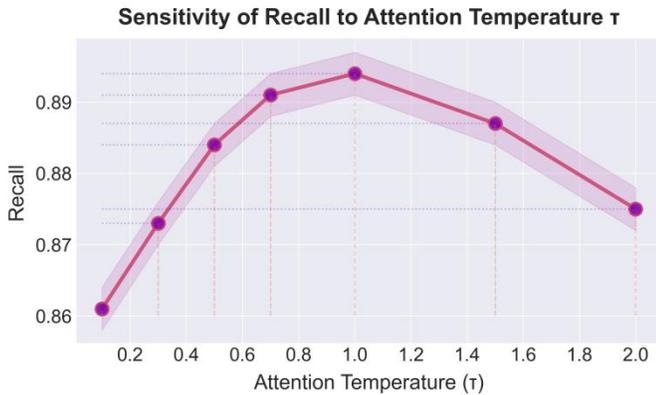

Figure 2. Experiment on the sensitivity of attention temperature $\tau$ to recall

As shown in Figure 2, the attention temperature parameter $\tau$ exhibits a clear sensitivity trend with respect to the model's Recall. When $\tau$ is small, the Recall is relatively low, indicating that overly strong attention concentration causes the model to focus excessively on local features while neglecting the global temporal dependencies within the EHR sequence. In this case, the model's perception of key clinical signals is limited, making it difficult to capture latent patterns across multiple temporal scales. As a result, some high-risk samples are not effectively identified.

As $\tau$ gradually increases, the Recall shows a significant upward trend and reaches its peak around $\tau \approx 1.0$. This suggests that a moderate attention temperature balances the model's focus between local and global features. It enables the temporal alignment module to perform dynamic weighting across different time scales, thereby enhancing the model's temporal robustness and risk recognition capability under asynchronously sampled EHR data.

When $\tau$ continues to increase, the Recall starts to decline slowly, indicating that an excessively high temperature leads to overly smooth attention distributions. This reduces the model's ability to distinguish between critical time points. Over-averaging of temporal weights weakens the selectivity of multi-scale temporal modeling, making it harder for the model to focus on key clinical moments and decreasing the accuracy of high-risk event identification. This trend shows that the temporal alignment mechanism in EHR modeling requires fine adjustment to maintain a proper balance between global dependency and local feature representation. A properly chosen $\tau$ not only improves the model's representation capability for complex medical time series but also maintains stable risk identification performance under non-uniform sampling and asynchronous events. This finding provides a useful reference for parameter optimization in clinical applications and further demonstrates the crucial role of multi-scale temporal modeling and adaptive alignment in improving medical risk prediction accuracy. This paper also presents an experiment on the sensitivity of the maximum sequence length limit L_max to F1-Score, and the experimental results are shown in Figure 3.

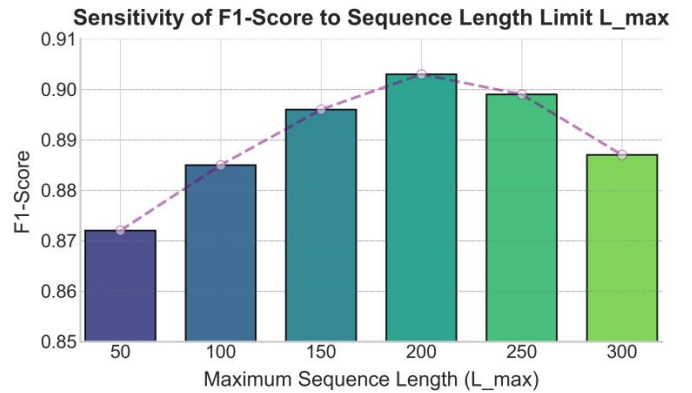

Figure 3. Sensitivity experiment of the maximum sequence length limit L_max to F1-Score

As shown in Figure 3, the maximum sequence length limit (Lmax) exerts a nonlinear influence on the model's F1-Score. When Lmax is small, the model fails to capture long-term dependencies in patients' EHR data, learning only local features and missing broader diagnostic trajectories, which lowers predictive accuracy. As Lmax increases, performance improves and peaks around 200, indicating that an appropriately extended sequence enables the model to align multi-scale temporal features and capture both short-term fluctuations and long-term patterns. However, beyond Lmax = 250, performance declines due to noise, redundancy, and increased computational burden that blur attention focus and weaken sensitivity to risk events. Overall, the results highlight that a well-chosen sequence length balances long-term dependency modeling with efficiency, ensuring optimal trade-offs between information richness and computational stability in EHR-based risk prediction.

## V. CONCLUSION

This study addresses the challenges of temporal asynchrony, uneven sampling, and multi-scale dependencies in Electronic Health Records (EHR) by proposing a risk prediction method based on a multi-scale temporal alignment network. The method employs hierarchical temporal modeling and adaptive alignment mechanisms to effectively capture the dynamic evolution of patient states under irregular temporal conditions, achieving unified representation across multiple time granularities. Experimental results show that the proposed model significantly improves performance in metrics such as accuracy, recall, and F1-Score, indicating its ability to extract more discriminative temporal features from complex medical sequences. From a structural perspective, the study validates the effectiveness of multi-scale temporal modeling and alignment mechanisms, providing a new technical pathway for clinical risk identification and a theoretical foundation for time-series modeling in EHR analysis.

Future work will further explore the adaptability and interpretability of the proposed model in cross-institutional and multimodal medical data. On one hand, integrating diverse data sources such as medical imaging, genomic information, and clinical text can help build a unified multimodal temporal fusion framework, enhancing the model's understanding of complex disease patterns. On the other hand, the interpretability of the model needs to be strengthened. Incorporating attention visualization and causal inference mechanisms can provide more transparent evidence for clinical decision-making. In the long term, the proposed multi-scale temporal alignment framework has the potential to be applied in broader medical and public health contexts, such as chronic disease management, disease progression prediction, and personalized treatment recommendation. It provides theoretical support and practical value for the development of intelligent healthcare systems and the implementation of precision medicine.